\documentclass{article}
\usepackage{spconf,amsmath,graphicx}
\usepackage[colorinlistoftodos]{todonotes}
\usepackage{url}

\title{LSTM-BASED WHISPER DETECTION}
\name{%
\begin{tabular}{@{}c@{}}
Zeynab Raeesy$^{\dagger}$ \qquad 
Kellen Gillespie$^{\dagger}$ \qquad
Zhenpei Yang$^{\ddagger}$\sthanks{The author contributed to this work during internship at Amazon Alexa}\\ 
Chengyuan Ma$^{\dagger }$ \qquad 
Thomas Drugman$^{\dagger}$ \qquad 
Jiacheng Gu$^{\dagger}$ \\
Roland Maas$^{\dagger}$ \qquad 
Ariya Rastrow$^{\dagger}$ \qquad 
Bj{\"o}rn Hoffmeister$^{\dagger}$ 
\end{tabular}}
 \address{$^{\dagger}$ Amazon Alexa \\
     $^{\ddagger}$University of Texas at Austin}

\begin{document}
%
\maketitle
\begin{abstract}
This article presents a whisper speech detector in the far-field domain. The proposed system consists of a long-short term memory (LSTM) neural network trained on log-filterbank energy (LFBE) acoustic features. This model is trained and evaluated on recordings of human interactions with voice-controlled, far-field devices in whisper and normal phonation modes. We compare multiple inference approaches for utterance-level classification by examining trajectories of the LSTM posteriors. In addition, we engineer a set of features based on the signal characteristics inherent to whisper speech, and evaluate their effectiveness in further separating whisper from normal speech. A benchmarking of these features using multilayer perceptrons (MLP) and LSTMs suggests that the proposed features, in combination with LFBE features, can help us further improve our classifiers. We prove that, with enough data, the LSTM model is indeed as capable of learning whisper characteristics from LFBE features alone compared to a simpler MLP model that uses both LFBE and features engineered for separating whisper and normal speech. In addition, we prove that the LSTM classifiers accuracy can be further improved with the incorporation of the proposed engineered features.\end{abstract}
\section{Introduction}
\label{sec:intro}

Advancements in speech and language technologies have resulted in the deployment of dialogue systems in real environments. These environments range from noisy living rooms with background noise to rooms that are very quiet. In the latter type of environment, a user may wish to whisper to the device, and in return would expect a response in a quieter and/or whispered voice. Triggering the suitable response mode on the device first requires detecting the whisper speech of the user. While automatic speech technologies are well researched and evaluated on normal speech in various acoustic conditions, there has been little effort in developing such technologies for other types of phonated speech such as whisper.

Whisper speech is mainly characterized as unvoiced speech due to a lack of periodic excitation in vocal folds. Studies of spectrograms have suggested that whisper speech overall has less energy at lower frequency bands compared to normal speech \cite{wenndt2002}. In \cite{jovicic1998} and \cite{wilson1985}, a consistent increase in F1 formant frequency in whisper speech in comparison to normal speech was reported. The signal characteristic differences between whisper and normal phonation have been the basis for several studies on classifying the two modes. Wenndt et. al \cite{wenndt2002} proposed a classification approach on whisper versus normal phonation using the energy ratios between high-frequency and low-frequency bands. In a study by Zhang and Hansen \cite{zhang2007}, several speaking modes such as whisper, soft, normal, loud, and shouted were investigated. A GMM-based classification system was developed based on sound intensity level, sentence duration, silence percentage, frame-energy distribution, and spectral tilt for these five categories of phonation. Spectral information entropy (SIE) was estimated in \cite{zhang2008} from probability density functions calculated over all frequency components, and the energy ratios between SIEs of multiple frequency sub-bands were used to form a 9-dimensional feature vector used in training GMM-based vocal effect classifier. The insensitivity of these features to absolute energy values in turn makes the classifiers built with such features robust to varying energy levels in input signals. This work was further enhanced in \cite{zhang2009} by using improved features with more frequency sub-bands for calculating SIEs.

To the best of our knowledge, the majority of the work in the related literature focuses on detecting and devising  relevant features for classifying whisper and normal speech. 
Compared to traditional machine learning models, deep and recurrent neural network models are capable of learning complex features even from raw data, with less reliance on task specific engineered features. One such example is voice activity detection (VAD), where DNN-based approaches \cite{hughes2013,maas2016} have proven superior in performance over classic approaches developed around engineered features \cite{shin2000,kristjansson2005}, such as audio energy \cite{woo2000}, pitch \cite{chengalvarayan1999}, zero-crossing rate \cite{benyassine1997,lu2002}, and cortical features \cite{thomas2012}.

In this paper, we propose using long-short term memory (LSTM) neural network models for the task of whisper detection. LSTM networks have proven to be successful classifiers in ASR, having been applied to a variety of tasks such as acoustic modeling \cite{sak2014} and endpoint detection \cite{maas2017}. We use a dataset of real recordings of natural human interactions, in whisper and normal speech, with a far-field voice-controlled speaker. To the best of our knowledge, this is the first work reporting on application of deep neural networks to whisper speech detection. Comparing to a baseline simple multilayer perceptron, LSTMs are proven to achieve a significantly higher frame accuracy. Based on our observation of LSTM posterior trajectories, we examine a number of inference modules for classification of utterances. In addition, inspired by the literature on whisper classification, we study the application of engineered features to this task. We examine classification performance with simple multilayer perceptron (MLP) models and LSTMs by adding a 6-dimensional vector of engineered features that are useful in distinguishing whisper/normal speech and compare them with models trained only on LFBE features. Based on our findings we prove that, in addition to scaling better, a more complex model such as an LSTM can perform reasonably well without the computational burden of engineered features. Through experiments and evaluations, we show that LSTM's performance can be further improved using the additional engineered features.

This paper is organized as follows. In the first part, we present an overview of the proposed classifier and inference mechanisms in sections \ref{sec:final-classifiers} and \ref{sec:inference}. Experiments and evaluations in section \ref{sec:experiments} comprises the dataset specifications in \ref{sec:data-preparation}, metrics in \ref{sec:metrics}, evaluation results in \ref{sec:evaluation}, and inference comparisons in \ref{sec:decision-making}. In the second part, sections \ref{sec:whisper-features} and \ref{sec:benchmarking} introduce the engineered features for whisper detection and evaluate the effectiveness of the features in classification task. Finally, the conclusions of this work and plans for future work are provided in section \ref{sec:conclusion}.

\section{LSTM-based Whisper Detector}
\label{sec:LSTM}

In this section, we begin with an overview of LSTM neural networks and their application to our whisper detector. We then discuss how we perform inference for utterance-level decision making from the frame-level posteriors of the LSTM classifier.

\subsection{Overview of the Classifier}\label{sec:final-classifiers}

The input data to the whisper classifier is in the form of sequential frames. Standard feed-forward MLP networks, with no concept of memory, do not allow us to use this data in an intuitive sequential, contextual way. Recurrent neural networks (RNNs) use feedbacks from their internal states in processing sequences of inputs, and thus consider the history of their states when modeling sequential data. However, RNNs are limited to short-term memory, as they suffer from the vanishing/exploding gradient problem \cite{bengio1994}. Long short-term memory (LSTM) models are extensions of RNNs, where memory cells with input, output, and forget gates are introduced at each recurrent layer to control the flow of information, consequently facilitating the learning of both short and long term dependencies in the input sequences \cite{hochreiter1997}. 

For the whisper classifier, LSTM models are trained using sequences of frames and their labels. Since this application of the model requires utterance-level decisions, each utterance in our dataset is tagged as whisper/non-whisper. These tags are propagated as target labels to all frames of that particular utterance. The model is trained using a cross-entropy objective function and is optimized with stochastic gradient descent (SGD) \cite{bottou1991} using the backpropagation through time (BPTT) algorithm \cite{williams1990}.

\subsection{Inference}\label{sec:inference}

The whisper classification models are structured to output scores at the frame level. Given a set of individual frame scores across a given utterance, we must then use an inference module, or result building process, to generate a classification score at the utterance level.

\begin{figure}[ht]
  \centering
  \includegraphics[width=1\linewidth]{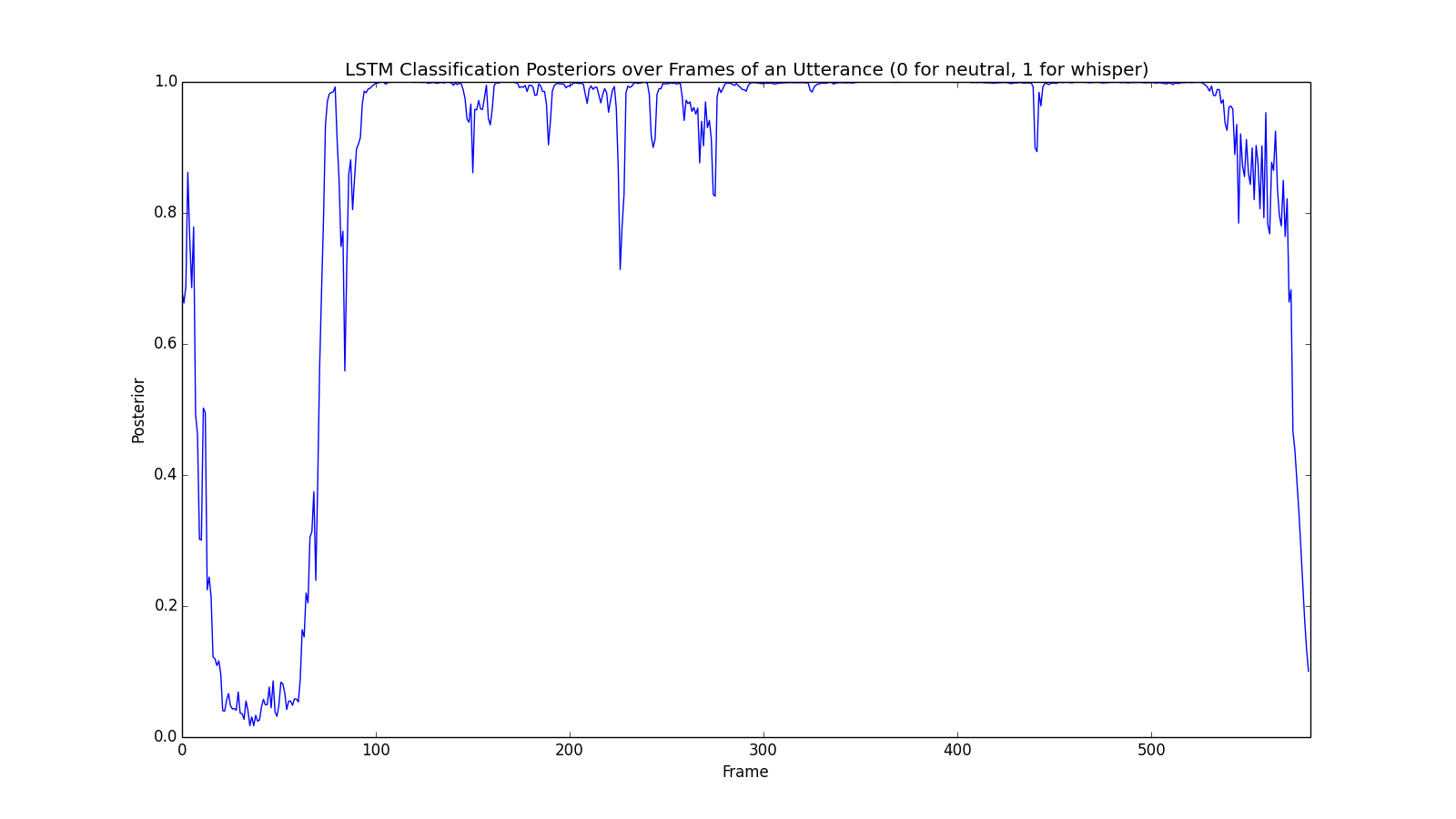}
  \caption{LSTM classification posteriors over frames of an utterance}\label{fig:posteriors}
\end{figure}

Upon performing posterior analysis of our model predictions on whisper (positive) test cases, we often observe sharp drops in posterior values towards the final frames of utterances. With the \textit{last-frame} inference module, these drops in turn result in sudden changes in utterance level predictions. An example of this behavior is shown in Figure \ref{fig:posteriors}. After investigating the audio, we find that these drops generally coincided with short trails of silent or near-silent frames found at the end of utterances. In many of these cases, as shown by the aforementioned figure, the model is confident in predicting whisper for long periods of time, only to fall sharply in the final frames.

To better represent the model predictions over the course of an entire utterance, we experiment with simple alternative inference modules. There are three main `classes' of inference module investigated, explained below.

\textbf{\textit{last-frame}}: Takes the last frame posterior.

\textbf{\textit{window-N}}: Takes the mean posterior of a window of the last $N$ frames.

\textbf{\textit{mean}}: Takes the mean posterior of all frames.

In addition, we investigate applying an offset to computing the average over the windows. We append the module name with \textit{ignore-last-50} in these cases. The 50 frames offset is based on trailing silence lengths previously observed by the production end-of-utterance detector on live traffic data, and was utilized to explore whether LSTM posteriors are more strictly tied to end of speech rather than end of the utterance.
Overall, we find that the inference modules that consider more frames outperform \textit{last-frame}, notably in terms of recall. The full results of the investigation are provided in section \ref{sec:decision-making}.

\section{Experiments and Evaluations} \label{sec:experiments}

\subsection{Data Preparation}\label{sec:data-preparation}

A series of whisper and normal utterances in US English was recorded in a quiet environment setup. We refer to this test set as ``in-house'' in the rest of this paper. The microphones used for collection were each placed between 1--2 feet away from the speaker. With the exception of 3 recording sessions that had fan noise, the audio was generally recorded in clean conditions with no background noise. In addition, we incorporated data from real recordings of natural human interactions with voice-controlled far-field devices containing only normal speech. In both dataset categories, the audio sampling rate was 16kHz, and for each utterance, the target label (whisper or normal) was propagated to all frames in that utterance.

From the in-house dataset of roughly 28k utterances, around 23k utterances were used in training and cross-validation, and the remaining 5k were selected for testing purposes, with no speaker overlap between train and test sets. The in-house test set consists of 3670 whisper and 1565 normal utterances. From the real recordings set, we added 30k utterances for training and cross-validation, and withheld 11k utterances for evaluating the false-positive rate on normal speech.We refer to the former as ``live traffic train'' set and to the latter as ``live traffic test" set.

\begin{table*}[ht]
\begin{center}
 \caption{Comparison of LSTM and MLP trained on LFBE features on in-house and live traffic tests.}\label{tbl:results}
  \begin{tabular}{ | c | c | c | c | c | c | c |}
  \hline
  model & feature & \multicolumn{3}{c}{in-house} & \multicolumn{2}{c|}{live traffic tests}  \\  
  \cline{3-7} 
          &            &  frame acc. & FPR (OP) & recall & frame acc. & FPR\\ \hline \hline
 MLP & LFBE   & 77.1\% &  0.1\% & 95.1\% & 94.9\% &  1.51\% \\ \hline
 LSTM & LFBE & 93.5\% & 0.1\%  & 97.4\% & 99.8\% &  0.21\% \\ \hline
  \end{tabular}
\end{center}
\end{table*}
\subsection{Metrics}\label{sec:metrics}

The ultimate goal of the whisper detector is to decide if an input speech signal is whisper or normal. Thus, in addition to raw frame accuracy accumulated from LSTM posteriors, we use recall and false-positive rate metrics at the utterance level. The frame accuracy is calculated at the default threshold of 0.5 with no tuning. To have meaningful comparison of the models, we tune the model thresholds to achieve 0.1\% false-positive rate on in-house tests. The tuned operating point (OP) is then used to compare the models in terms of false-positive rate on live traffic test set and recall on in-house sets. To have the overall picture, we also compare the models in terms of F1 score on accumulation of test sets.

\subsection{Classifier Evaluation}\label{sec:evaluation}

We extract 64-dimensional LFBE features for every 25ms frame of the utterance, with 10ms overlap between the frames. Channel mean subtraction (CMS) is applied to utterances on a per-speaker, per-device basis in real recordings    and per-speaker in in-house test data. The LSTM model structure consists of 2 hidden layers each with 64 memory cells. The output layer is 2-dimensional, corresponding to whisper and normal status. The baseline system is a simple multilayer perceptron (MLP) with 3 hidden-layers and a 2-dimensional output layer. 

The final utterance level results are built using the mean of the frame posteriors of the entire utterance. This inference module was chosen empirically based on experiments explained in section \ref{sec:decision-making}.

Table \ref{tbl:results} shows frame accuracy comparisons of the LSTM and MLP model on in-house and live traffic test sets. For fair comparison of  models, the threshold has been tuned for both models to have equal false-positive rate on in-house test sets (the operating point OP). As observed, the LSTM outperforms the MLPs on both in-house recall and live traffic test sets FPR at false-positive rate of 0.1\% on in-house. However a larger gap is observed on in-house test set which contains both whisper and non-whisper utterances, suggesting both positive and negative instances are contributing to the difference in the classification outcomes of the two models. 

\subsection{Inference at Utterance Level}\label{sec:decision-making}
Table \ref{tbl:resultbuilders} shows the results of our inference module investigation. As expected, the inference modules that consider more frames have improved recall over the \textit{last-frame} module. Our false-positive rate over live traffic test set examples does increase for some modules, with varying magnitudes. Overall, the simple \textit{mean} module proved best on this test set.

\begin{table*}[h]
\begin{center}
 \caption{Comparison of Result Building Modules using posteriors of LSTM}\label{tbl:resultbuilders}
  \begin{tabular}{| c | c | c | c | c |}
    \hline
    result builder & \multicolumn{2}{c|}{FPR} & recall & F1 score \\
    \cline{2-3} \cline{4-5}
    & in-house & live traffic test set & in-house & all \\ \hline
    \hline
    last-frame				&	0.1\%	&	0.1\%	&	94.1\% &	96.7\% \\ \hline
    window-100-ignore-last-50	&	0.1\%	&	1.6\%	&	95.1\% &	94.3\% \\ \hline
    mean-ignore-last-50		&	0.1\%	&	0.4\%	&	97.1\% &	97.7\%\\ \hline
    mean					& 	0.1\%	&	0.2\%	&	97.4\% &	98.2\%\\ \hline
  \end{tabular}
\end{center}
\end{table*}

\section{Feature Study}\label{sec:features}

\subsection{Classifier Features}\label{sec:whisper-features}

Three categories of features are studied in this work: sum of residual harmonics (SRH), high-frequency energy (HFE), and features based on auto-correlation of time-domain signal (ACMAX). A review of these features is presented below.
\newline

\textbf{\textit{Sum of Residual Harmonics (SRH):}}
Whisper speech is typically characterized by the absence of fundamental frequency (F0) due to a lack of voicing. The SRH feature, originally proposed for robust pitch tracking in noisy conditions by Drugman and Alwan \cite{drugman2011}, is used as a voicing detector in this work. The SRH feature uses harmonic information in the residual signal and is calculated as:

\begin{equation}
SRH(f)= E(f) + \sum_{k=2}^{N_{harm}} [E(k \cdot f ) - E((k-\frac{1}{2}) \cdot f)]
\end{equation}

\noindent where $E(f )$ is the amplitude spectrum for each Hanning-windowed frame, and for voiced speech presents peaks at the harmonics of F0. The second term in summation, \(E((k-\frac{1}{2}) \cdot f))\), helps reduce the relative importance of the maxima of SRH at even harmonics. The value of SRH is sensitive to the initial FFT size, and higher FFT sizes lead to better separation between the values of SRH features in whisper versus normal speech.

\textbf{\textit{High Frequency Energy (HFE):}}
Inspired by the observations in \cite{wenndt2002} about power spectrum differences in low/high band frequencies between whisper and normal speech, the HFE feature consists of two dimensions. The first dimension reflects the energy ratio between the high frequency band (6875\url{~}8000hz) energy and the low frequency band (310\url{~}620hz) energy. Whisper generally has less energy in lower frequency bands, thus this ratio can be effective in distinguishing whisper and normal speech. The high and low frequency bands are empirically selected to maximize the separation. The second dimension is the Shannon entropy of the low frequency area. This entropy is calculated by treating the power spectrum as a probability distribution. Whisper tends to have high entropy in the low frequency band.

\textbf{\textit{Auto-Correlation Peak Maximum (ACMAX):}}
The maximum autocorrelation peak within the plausible human F0 range (80\url{~}450 hz) is calculated and used as the first dimension for this feature. A value is identified as a peak if it is larger than its 4 neighbors on the left and right. The second and third dimensions of the ACMAX feature consist of the position of the peak and the mean distance between consecutive autocorrelation peaks, respectively. 

\color{black}
\subsection{Feature and Classifier Evaluation\label{sec:benchmarking}}

To evaluate the effectiveness of the features, we trained our LSTM and MLP models with new input vectors consisting of 64 LFBE features plus the 6 engineered features discussed in previous paragraphs. Table \ref{tbl:features} shows the evaluation results on in-house and live traffic test sets.

\begin{table*}[t]
\begin{center}
 \caption{Comparison of classifiers trained on LFBE only and LFBE + engineered features.}\label{tbl:features}
  \begin{tabular}{| c  | c | c | c | c | c | c | c |}
    \hline
    model & input features &  \multicolumn{3}{c|}{in-house test set} & \multicolumn{2}{c|}{live traffic test set} & \\ \hline 
    \cline{3-5} \cline{6-7}
   & & frame acc. & FPR (OP) & recall & frame acc. & FPR & F1 score\\ \hline \hline
   MLP & LFBE & 77.1\% & 0.1\% & 95.9\% & 94.9\% & 1.5\% & 94.9\% \\\hline
   LSTM & LFBE & 93.5\% & 0.1\% &  97.4\% & 99.7\% & 0.2\% & 98.2\% \\ \hline
   MLP & LFBE + engineered  &74.6\% &	0.1\% & 98.8\%	& 98.0\%	& 0.6\%  & 98.2\% \\ \hline
   LSTM & LFBE + engineered  & 96.0\% &	0.1\% & 99.3\% & 99.9\% & 0.1\% & 99.6\% \\ \hline
  \end{tabular}
\end{center}
\end{table*}

The addition of the engineered features to the existing LFBE features improves both models. For the LSTM model, the engineered features help improve the frame accuracy by 3.5\%, leading to more than 99\% recall on in-house whisper utterances while reducing rate of false-positives by half. We further observe the LSTM trained only on LFBE features performs comparably to the MLP trained with LFBE and engineered features. This observation suggests that the LSTM model can indeed learn more of the underlying characteristic differences of whisper speech from LFBE features in comparison with the MLP model. While the recall values for the LFBE LSTM are slightly lower than the LFBE + engineered MLP in this case, an improvement in false-positive rate is observed in comparison and the models share similar F1 scores at the same operating point.

For the MLP model, despite the frame accuracy drop in the MLP trained on LFBE + engineered features, the engineered features improve the model performance in terms of in-house recall, live traffic FPR, and overall F1 score. The drop in MLP frame accuracy could potentially be attributed to a caveat in our data labels where the utterance level tags, i.e. whisper/non-whisper, are propagated to all the frames. This includes both speech frames of interest and non-speech frames such as silence and non-speech noise. In reality, the silence and noise frames need to be labeled separately as they are common and indistinguishable between whisper and non-whisper utterances. The addition of engineered features in the MLP, while not addressing the confusion at the frame-level, seems to be helping to address this issue at the utterance level. 

\section{Conclusions}\label{sec:conclusion}

In this work, we proposed using LSTM networks for the task of detecting whisper speech using standard and widely-used LFBE features. We also developed and reviewed a set of features engineered for the task of whisper speech classification and compared the detection ability of the models with and without these engineered features. 
Our findings show that, with sufficient data, LSTMs can learn the underlying characteristic differences of whisper speech from LFBE features alone, without requiring more sophisticated engineered features. This representational power with standard features makes these LSTMs better candidates for large-scale applications. We show we can further improve the LSTM model performance by utilizing the engineered features in addition to the original LFBE features.

In future work, we plan to experiment with more complex and informed inference modules, including a module using an underlying voice activity detection (VAD) model to filter or weight frames based on their likelihood of containing speech content. We also plan to improve our models' robustness to varied recording conditions and languages by incorporating mixed-condition and mixed-language data into our training and evaluation.

\bibliography{SLT-whisper} 
\bibliographystyle{IEEEbib}

\end{document}